%% file: bmvc_review.tex
\documentclass{bmvc2k}

\usepackage[utf8]{inputenc} % allow utf-8 input
\usepackage[T1]{fontenc}    % use 8-bit T1 fonts
\usepackage{hyperref}       % hyperlinks
\usepackage{url}            % simple URL typesetting
\usepackage{booktabs}       % professional-quality tables
\usepackage{amsfonts}       % blackboard math symbols
\usepackage{nicefrac}       % compact symbols for 1/2, etc.
\usepackage{microtype}      % microtypography
\usepackage{xcolor}         % colors
\usepackage{bm}

\usepackage{graphicx}
\usepackage{amsmath}
\usepackage{amssymb}
\usepackage{subcaption}
\usepackage[capitalise]{cleveref}
\usepackage{rotating}

\include{macrosjvg}

\title{Learning to Adapt to Position Bias in Vision Transformer Classifiers}

\addauthor{Robert-Jan Bruintjes}{http://rjbruintjes.nl}{1}
\addauthor{Jan van Gemert}{https://jvgemert.github.io/}{1}

\addinstitution{
 Computer Vision lab\\
 Delft University of Technology\\
 Delft, the Netherlands
}

\runninghead{Bruintjes \& van Gemert}{Learning to Adapt to Position Bias in Vision...}

\def\eg{\emph{e.g}\bmvaOneDot}

\begin{document}

\makeatletter
\@ifundefined{thesubfigure}{
  \PackageWarningNoLine{subcaption}{Subfigure support may not be working properly!}
}{}
\makeatother

\maketitle

\begin{abstract}
\input{sections/0_abstract}
\end{abstract}

%-------------------------------------------------------------------------
\input{sections/1_introduction}
\input{sections/2_related_work}
\input{sections/3_method}
\input{sections/4_experiments}
\input{sections/5_discussion}

\bibliography{bib}

\newpage
\appendix

\input{sections/9_appendix}

\end{document}

%% file: macrosjvg.tex
\usepackage{xcolor}

\usepackage{pifont}
\usepackage{caption}
\usepackage{tabularx} % extra features for tabular environment
\newcolumntype{Y}{>{\centering\arraybackslash}X}

\usepackage{booktabs}
\usepackage{multirow}

\newcommand{\bX}{\mathbf{X}}
\newcommand{\bP}{\mathbf{P}}
\newcommand{\pe}{\mathbf{P}}
\newcommand{\pshap}{\text{P-SHAP}}

\def\x{{\boldsymbol x}}
\def\W{{\boldsymbol W}}

%% file: sections/0_abstract.tex
How discriminative position information is for image classification depends on the data. On the one hand, the camera position is arbitrary and objects can appear anywhere in the image, arguing for translation invariance.
At the same time, position information is key for exploiting capture/center bias, and scene layout, \eg: the sky is up.
We show that \textit{position bias}, the level to which a dataset is more easily solved when positional information on input features is used, plays a crucial role in the performance of Vision Transformers image classifiers. 
To investigate, we propose \textbf{Position-SHAP}, a direct measure of position bias by extending SHAP to work with position embeddings.
We show various levels of position bias in different datasets, and find that the optimal choice of position embedding depends on the position bias apparent in the dataset. We therefore propose \textbf{Auto-PE}, a single-parameter position embedding extension, which allows the position embedding to modulate its norm, enabling the unlearning of position information. Auto-PE combines with existing PEs to match or improve accuracy on classification datasets. 

%% file: sections/1_introduction.tex
\section{Introduction}
\label{sec:intro}

%%% ViTs for classification should not learn position and should learn position -> how does position affect ViTs for classification?
Vision Transformers~\cite{dosovitskiy2021imageworth16x16words} explicitly infuse their learned representations with information on the position of features using position embeddings.
This goes against the long-held belief that, when trained specifically for image classification, the learned representations of vision models like Convolutional Neural Networks should be free of positional information (e.g. be translation invariant)~\cite{biscione2021convolutionalneuralnetworksinvariant,kayhan2020translation,zhang2019makingconvolutionalnetworksshiftinvariant,islam2021globalpoolingmeetseye}.
However, adding position embeddings often improves the performance of Vision Transformers significantly in image classification~\cite{akkaya2023enhancingperformancevisiontransformers,cordonnier2020relationshipselfattentionconvolutionallayers,dosovitskiy2021imageworth16x16words,d_Ascoli_2022,Raghu2021,Ren_2023_CVPR}.
% showing that Vision Transformers need position information to solve image classification well
This raises the questions \textit{how} and \textit{why} position information affects Vision Transformers.
This work investigates these questions.

\begin{figure}
\centering
\includegraphics[width=1.0\linewidth]{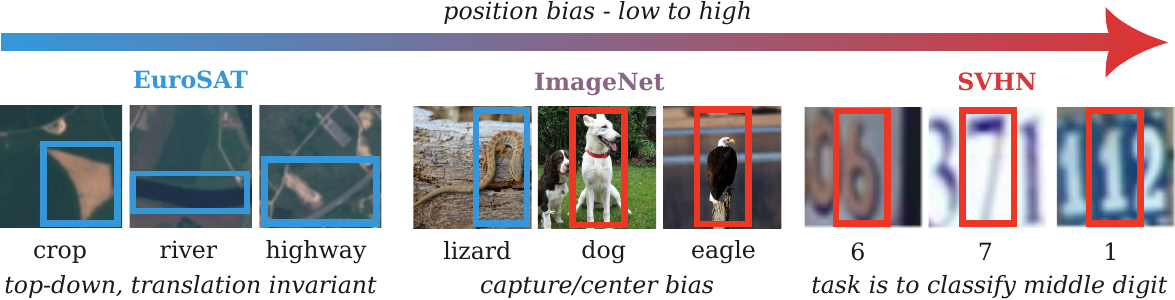}
\vspace{1mm}
\caption{
In addition to appearance information, some classification datasets require position information to be solved, and some do not. In EuroSAT, the object of interest appears anywhere in the pecture. In ImageNet, often but not always the object of interest is centered in the frame ("capture bias"). In SVHN, the task is explicitly to classify only the middle digit.
% With only appearance information, the classifier may not give the correct response. The network needs to use the position embedding to prioritize classifying the race of the middle dog only.
% Move SVHN also to fig 1? Fig 1 now has "too much" info, the blue boxes and arrows are not needed? (also not mentioned in the caption). Make fig 1 about showing various kinds of position biases, (not about VITs: too technical, whereas fig 1 should be 'motivational') }.
% \jvg{I would make this fig about that the importance of position information is dataset specific. So, show examples of datasets with different position importance; maybe even already with the estimated importance by P-Shap? }
}
\label{fig:figure-1}
\end{figure}

% Position bias in various datasets means we need to measure it
We define \textit{position bias} in datasets as the level to which knowing the spatial position of objects of interest in the input images of the dataset helps the model to learn to classify the data. An example of a particularly common source of position bias is \textit{capture bias}~\cite{Fabbrizzi2022,TorralbaEfros2011}, which occurs when the object of interest is commonly framed in the center of the photograph.
When multiple objects are present, the model should learn to only classify the center object, and therefore needs to use information on the position of input features (see Fig.~\ref{fig:figure-1}).
% ImageNet and other datasets of natural images can be assumed to have capture bias.
Other position biases are even more overt, such as the bias apparent in the SVHN~\cite{Netzer2011} dataset, where the classification task is explicitly to only classify the center digit in the image.
There are also datasets that theoretically do not contain any position bias, such as EuroSAT~\cite{helber2018introducing,helber2019eurosat}, which is comprised of satellite imagery where the camera position is independent of the content.
We conclude that the level of position bias in classification datasets varies, and hypothesize that the ability of classification models to model this position bias affects the accuracy of these models.

In this work, we explore the effects of varying position bias in classification datasets on the accuracy of Vision Transformers. First, to know the level of position bias in a dataset, we should measure it. However, no direct measurement method exists to measure position bias.
Therefore, we propose \textbf{Position-SHAP} (P-SHAP), a direct measure of position bias learned by classification models.
% , by extending SHAP to treat position embeddings as input features.

% Therefore, we desire a direct measure of position bias to reveal the optimal position embedding from a single training run on a baseline model, instead of tuning the model extensively on various position embeddings.
% However, no methods exist for directly measuring position bias. 
% existing works either require special experimental setups, including modifying data and/or models, to estimate (notions similar to) position bias~\cite{Kayhan_2020_CVPR,zhang2019makingconvolutionalnetworksshiftinvariant}.
% , or merely improve accuracy by changing the way the network deals with position information, without measuring position bias itself~\cite{caron2023locationawareselfsupervisedtransformerssemantic,d_Ascoli_2022,Camporese2022WhereAM,kong2022spvitenablingfastervision,liu2021efficient,liu2021swintransformerhierarchicalvision}.

%%% Experimental questions
% We verify that P-SHAP measures position bias in a controlled CIFAR-derived setting.

Using P-SHAP, we measure position bias in models trained on several classification datasets and find that models trained on datasets with high position bias do better when learning more position bias.
Furthermore, we show that using position-affecting augmentations (such as RandAugment~\cite{cubuk2020randaugment}) and transfer learning affect the learned position bias.
% We find that transfer learning affects position bias, slightly reducing it. Datasets with high position bias lose more accuracy by leaving out fine-tuning than datasets with low position bias.
% We find that position bias measured in trained models matches hypothesized position biases on the EuroSAT and SVHN datasets.

Finally, we show that knowing the position bias of a dataset informs the optimal choice of position embedding method. Specifically, we find that models trained on datasets with low position bias perform better without a position embedding than with a learnable absolute position embedding (APE). We therefore introduce a simple single-parameter extension to position embeddings that learns to reduce the norm of the position embedding, called \textbf{Auto-PE}. Adding Auto-PE to RoPE~\cite{heo2024rotarypositionembeddingvision} position embeddings matches or improves performance in all tested settings, without having to manually tune the method of position embedding.

All code and experiments are made publicly available at 
{\small\url{https://github.com/rjbruin/position-shap}}.
% {\small [URL withheld for anonymity]}.

%% file: sections/2_related_work.tex
\section{Related work}

\textbf{Vision Transformers and position embeddings.}
% \jvg{consider repacing subsections with textbf to save space. So, further use of textbf inside the subsection could become \emph{emph} or something else?}
Vision Transformers~\cite{dosovitskiy2021imageworth16x16words} apply transformers~\cite{vaswani2017attentionneed} to computer vision tasks. To make input images compatible with transformers, images are divided into flattened patches, whose dimensionalities are reduced by a linear mapping to match the hidden layers of the transformer, before position embeddings (PEs) are added:
% \jvg{do you use eq 1 elsewhere? If not, then consider doing eq 1 as inline; to save space}

\begin{align}
    \bX^P &= \text{Patchify}(\bX), \quad &\bX \in \mathbb{R}^{H \times W \times C}, \bX^P \in \mathbb{R}^{N_P \times (H_P \times W_P \times C)} \\
    \mathbf{Y}_i &= \text{Linear}(\bX^P_i) + \bP_i, \quad &i \in N_P, \pe \in \mathbb{R}^{N_P \times D}, \mathbf{Y} \in \mathbb{R}^{N_P \times D} \label{eq:vit}
\end{align}

where
$\bX$ is a single input image%
,
$H \times W \times C$ are the image dimensions%
,
$N_P$ is the number of patches%
,
$H_P \times W_P$ are the spatial dimensions of the patches%
,
$\bX^P$ are the patches of the input sample%
,
$\bP$ is the position embedding%
,
and
$\mathbf{Y}$ is the input to the first transformer block.
% The position embedding (PE) imbues the learned representation with information about the position of the input patch. 
Since transformers are permutation-invariant with respect to the patch dimension, learned representations cannot know their position without PEs.

There are many different PE methods.
We distinguish baseline methods, introduced during the initial development of ViTs, from original contribution methods, published in later years. 

% There are a number of position embedding methods for Vision Transformers: 
\textit{Baseline PEs.} The Absolute Position Embedding (APE)~\cite{vaswani2017attentionneed} learns a one-dimensional token for each input patch, which is initialized randomly. Several alternative methods initialize the APE using a fixed function (e.g. sinusoid-based~\cite{vaswani2017attentionneed}), and continue to either fine-tune the APE or keep it fixed during training. Alternatively, Relative Position Embedding (RPE)~\cite{shaw2018selfattentionrelativepositionrepresentations} methods infuse position information into the self-attention operation, by adding biases to the attention scores dependent on the relative position of tokens.

% In recent years, a number of more advanced PE methods have been proposed to improve the position bias, and therefore the performance, of ViTs.
\textit{Original contribution PEs.}
% Several extensions of the APE method have been proposed since the introduction of the Vision Transformer.
Learnable Fourier Features~\cite{li2021learnablefourierfeaturesmultidimensional} generalize the APE method initialized with sinusoidal features to multiple dimensions and learn the frequencies of the wave functions end-to-end.
CAPE~\cite{likhomanenko2021capeencodingrelativepositions} augments APE embeddings to reduce absolute positional information of features, while retaining relative positional information between features.
iRPE~\cite{wu2021rethinkingimprovingrelativeposition} extends the RPE method to allow for the bias term to be dependent on the input tokens, and groups relative distances in the computation of this bias according to their size.
Rotary Position Embedding (RoPE)~\cite{su2023roformerenhancedtransformerrotary,heo2024rotarypositionembeddingvision} integrates into the attention operation, like RPE, where it transforms incoming tokens to complex values by reshaping, before rotating the complex-valued input tokens around the origin by end-to-end learned angles, before transforming the tokens back to real values. Because the works proposing these models show improved accuracy, we analyze these PE methods for their effect on the position bias directly using P-SHAP.

\textbf{Position bias.}
% In the following, we discuss a number of works aimed at investigating or measuring position bias.
% \textit{Effects of position information in vision models.}
In the era before Vision Transformers, Convolutional Neural Networks (CNNs) dominated computer vision tasks. CNNs are designed to be translation invariant~\cite{lecun1998gradient} so as to share gradients among translated input features. As a result, they should not be able to model the position of input features. However, border effects~\cite{biscione2021convolutionalneuralnetworksinvariant,kayhan2020translation} and improper subsampling~\cite{zhang2019makingconvolutionalnetworksshiftinvariant,islam2021globalpoolingmeetseye} allow CNNs to learn positional information.
% , generally to their detriment. 
Several works aim to remove position information by improving translation invariance in CNNs~\cite{liu2018intriguingfailingconvolutionalneural,kayhan2020translation,zhang2019makingconvolutionalnetworksshiftinvariant} and in Vision Transformers~\cite{rojasgomez2023makingvisiontransformerstruly}, with improved performance as a result. However, adding position embeddings to Vision Transformers, thereby enabling models to learn positional information, improves performance as well. This work unifies these findings by showing that position information benefits performance on classification datasets that contain position bias, but does not benefit performance on datasets without position bias.

\textit{Related notions.} Various works have explored notions similar to position bias, coining terms such as "positional variation"~\cite{watters2019spatialbroadcastdecodersimple}, "structural learning"~\cite{jelassi2022visiontransformersprovablylearn}, "position difference" ~\cite{cordonnier2020relationshipselfattentionconvolutionallayers}, "local processing"~\cite{cordonnier2020relationshipselfattentionconvolutionallayers}, or "token mixing"~\cite{yu2022metaformeractuallyneedvision}. Some of these works also perform measurements on these notions, such as on locality (the distance between patch representations in the attention operation)~\cite{cordonnier2020relationshipselfattentionconvolutionallayers,Raghu2021}. However, any measure of something other than position bias may be affected by confounding factors. For example, using shifted pixels to measure change in accuracy, as in~\cite{zhang2019makingconvolutionalnetworksshiftinvariant}, can show the effect of position bias but is confounded by the effects of the patch size of the model, as shifting the image by a number of pixels equal to the patch size will yield identical patches.
We therefore propose a direct measure of position bias, unaffected by confounding factors.

% \paragraph{Perturbation analyses.} 
\textit{Testing performance.} Some works perform perturbation analysis and measure effects on accuracy~\cite{Kayhan_2020_CVPR,zhang2019makingconvolutionalnetworksshiftinvariant}.
% In contrast, we provide a method to directly quantify position bias. This allows us to analyze measurements for patterns that inform model designs.
Other works show improved performance by embedding inductive biases~\cite{d_Ascoli_2022,kong2022spvitenablingfastervision,liu2021swintransformerhierarchicalvision,ding2023revivingshiftequivariancevision} into models or using pretext tasks~\cite{caron2023locationawareselfsupervisedtransformerssemantic,Camporese2022WhereAM,liu2021efficient}, aiming to improve position bias. These results suggest position bias is important and should be investigated. However, these works do not measure position bias and can therefore not conclusively demonstrate that their improved results are due to position bias.
% \rj{Is this better? We do want to make clear why our contribution is important...?}

\textbf{Kernel SHAP and Vision Transformers.}
\label{sec:shap-vit}
Our measure of position bias builds on Kernel SHAP~\cite{LundbergLee2017}. Kernel SHAP is a feature attribution method, which means it assigns importance scores ("SHAP values") to each input feature in a single input sample $\bX$ of a neural network. Kernel SHAP assumes that the network response $f(\bX)$ can be approximated by a linear function $g(\bX)$:

\begin{align}
    \label{eq:additivity}
    % \hat{\mathbf{x}} &= \text{Flatten}(\bX) \in \mathbb{R}^{N_P} \\
    % f(\bX) \approx g(\bX) &= \phi^\bX_0 + \sum^{N^P+1}_{i=1} \phi^\bX_{i} \hat{\bx}_{i},
    f(\bX) \approx g(\bX) &= \phi^\bX_0 + \sum^{H}_{i} \sum^{W}_{j} \sum^{C}_{k} \phi^\bX_{ijk} \bX_{ijk} 
    % f(\bX) \approx g(\bX) &= \phi^\bX_0 + \sum^{H \times W \times C + 1}_{i=1} \phi^\bX_{i} \bX_{i},
\end{align}
where
% $\text{Flatten}()$ flattens $\bX$ to a vector $\bx$, and 
$\boldsymbol{\phi}^\bX \in \mathbb{R}^{H W C + 1}$ are the SHAP values that represent the impact of each input feature in $\bX$, with $\phi^\bX_0$ representing the impact of all collected biases of the network.

\textit{Estimating SHAP values.} The SHAP values $\boldsymbol{\phi}^\bX$ are estimated for each sample $\bX$ by weighted linear regression. The loss function depends on the difference between the actual network response $f(\bX)$, and $f$ applied to versions of $\bX$ where the values of all possible subsets of features are replaced with a \textit{background value}, to emulate removal of features. This background value is the mean value of the feature in a so-called \textit{background dataset}, which should have the same distribution as the sample, \eg the training dataset. See Appx.~\ref{sec:kernelshap} for details.
% For the sake of brevity, we refer to~\cite{LundbergLee2017} for rationale and details. \rj{Can I get away with this? I really don't want to go into details, for space as well as for how complex it makes the whole thing. If I can't, I'll put a more detailed explanation in the appendix...}

\textit{Kernel SHAP for PEs.} In this work we would like to use Kernel SHAP to estimate SHAP values for PEs. However, the values of the PE are the same for all samples in any background dataset, as they are a bias parameter and therefore constant w.r.t. the input (see Eq.~\ref{eq:vit} where $\bP$ does not depend on $\bX$). If the background value is equal to the value of the feature, the Kernel SHAP loss is zero valued, and the SHAP value cannot be estimated. Therefore, it is impossible to estimate SHAP values for the PE using Kernel SHAP as is. However, in this work, we propose Position-SHAP, an extension of Kernel SHAP that \textit{can} attribute SHAP values to position embeddings.

% \update{For details on the exact formulation of this optimization problem, see Appx.~\ref{sec:kernelshap}.}

%% file: sections/3_method.tex
\section{Position-SHAP}

% PEs in Vision Transformers are embeddings that are added to the patch embeddings.
% Therefore, they are not \update{weights but rather bias parameters} of the network. 
For Kernel SHAP to compute SHAP values for the PE, we need to choose a suitable way to set the background value of the PE.
To recap, for input features, the appropriate background dataset has the same distribution as the input feature. Our contribution is to consider each PE token $\bP_i \in \mathbb{R}^D$ in the PE to be a sample from a distribution of PE tokens learned by this model, and to consider the set of all PE tokens $\bP$ as the background dataset.
% We solve this problem by setting the background value of one PE token $\bP_i$ to the mean of the other PE tokens.
In practice, we extend the implementation of Kernel SHAP to include the PE as a feature by shuffling the PE along the token dimension when computing background values.
% Proof follows from additivity of Kernel SHAP: SHAP value for whole PE is the same as sum of SHAP values for each PE token

% \textbf{Grouping features.}
We are now able to compute SHAP values for each element of the PE. However, we would like to have a single SHAP value to represent the impact of the PE. Fortunately, because $g$ is a linear function, the SHAP value of a group of features is the sum of the SHAP values of the individual features. We can therefore compute a single SHAP value for the group of features.

Finally, \textbf{Position-SHAP} (P-SHAP) is defined as the ratio between the SHAP value of the PE and the SHAP value of the PE and image together:

% We can combine the SHAP values $\phi_i$ of features because SHAP values are \textit{additive}: the SHAP value of a group of two features is equal to the sum of the SHAP values of the individual features~\cite{LundbergLee2017}. In this way, we compute a single SHAP value $\phi_I$ for the entire input image $\bX$.

\begin{align}
    \label{eq:position-shap}
    % f(\bX) \approx g(\bX) &= \phi_0 + \sum^{N_P+1}_{j=1} \phi_j \bP_j \sum^{2 N_P + 1}_{i=N_p + 1} \phi_{i} \hat{\bx}_{i}, \\
    g(\bX) &= \phi^\bX_0 + \phi^\bX_P \sum \bP + \phi^\bX_I \sum \bX \\
    \text{Position-SHAP}(\bX) &= \frac{\phi^\bX_P}{\phi^\bX_P + \phi^\bX_I},
\end{align}
where $\phi^\bX_P$ is the SHAP value for the PE and $\phi^\bX_I$ is the SHAP value for the input image.

We compute P-SHAP for individual samples. In our experiments, the P-SHAP value of a model over a test set is taken as the average over the correctly classified samples in the test set. We note that Kernel SHAP (and therefore P-SHAP) has quite a high compute cost. We discuss additional hyperparameters as well as inference cost and consistency of P-SHAP in Appx.~\ref{sec:p-shap-hyperparameters}.

\section{Auto-PE}

% Inspired by our findings we propose several new simple position embedding extensions.

Inspired by our findings on the importance of position bias, we propose \textbf{Auto-PE}, a simple single-parameter method that learns to reduce or increase the impact of the position embedding by modulating the norm of the position embedding using a learned weight $\gamma$, before adding the position embedding to the patch embeddings:

\begin{align}
    \mathbf{Y}_i = \text{Linear}(\bX^P_i) + \underbrace{\gamma \times \bP_i}_{\text{Auto-PE}}.
    % \mathbf{Y}_i &= \text{Linear}(\bX^P_i) + \bP_i, \quad &i \in N_P, \pe \in \mathbb{R}^{N_P \times D}, \mathbf{Y} \in \mathbb{R}^{N_P \times D}
\end{align}

Why is Auto-PE necessary? Could an APE-like method with learnable values not just learn to decrease the norm of the PE by itself? We posit that it could, but that Auto-PE learns this solution much more effectively, because it is a classic example of an effective inductive bias: it captures the modulation of the PE norm in a single parameter, making it easier for the model to learn the modulation.

Auto-PE can be applied to any APE-based method that adds the position embedding to the patch embedding, by applying the gating to the computed position embedding. We initialize $\gamma = 0.5$ and optimize $\gamma$ end-to-end with a learning rate of $1e-1$. We tuned these hyperparameters on the test sets of EuroSAT and Flowers-102.

\textbf{Auto-RoPE.} We design a special case of Auto-PE for RoPE~\cite{heo2024rotarypositionembeddingvision} by applying $\gamma$ to the learned angles used in the complex-valued multiplication (i.e. rotation) that RoPE applies in the attention operation. In this way, when $\gamma$ approaches zero, the angles approach zero and rotation is reduced, reducing positional information.
% \update{We learn a single parameter $\gamma$ for the whole network, which means that $\gamma$ is shared between all attention operators in the network.}
We learn a single parameter $\gamma$ for each attention operator in the network.
See Appx.~\ref{sec:auto-rope} for details.

% \paragraph{SCAPE.} Skip-Connect APE adds skip-connections from the position embedding to the start of each Transformer block. In other words, the APE is added not only to the patch embeddings, but to the tokens before each attention block. We hypothesize this makes it easier for blocks deeper in the network to access position information and learn position bias.

% \paragraph{DRAPE.} Decay-Regularized APE adds weight decay regularization on the position embeddings. The motivation is to stimulate the model to reduce the norm of the position embeddings if they do not contribute to the network output. By default we use DRAPE with a magnitude of $1\mathrm{e}{-4}$.

%% file: sections/4_experiments.tex
\section{Experiments}

\subsection{Controlled setting}
\label{sec:type1}

\begin{figure}
\centering
\includegraphics[width=1.0\linewidth]{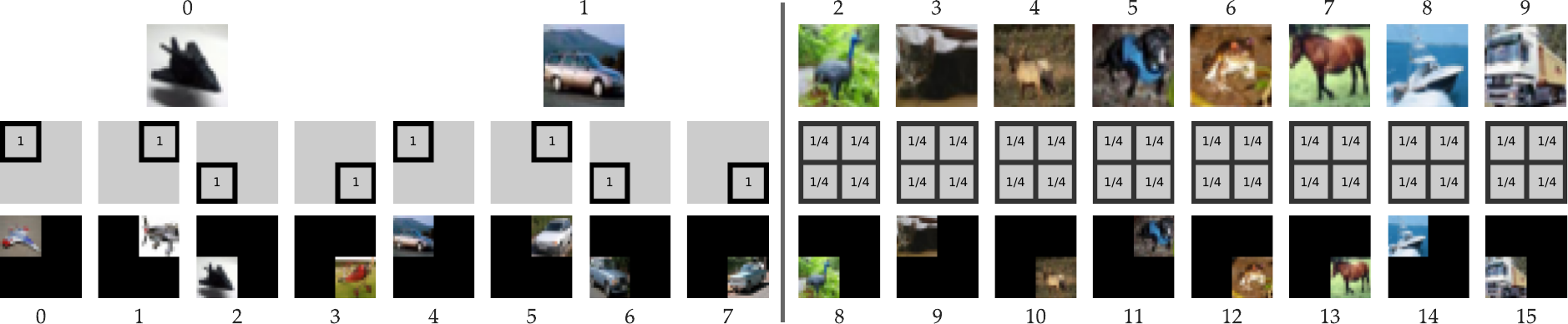}
\vspace{1mm}
\caption{
CIFAR-10-Position toy dataset construction. Top row shows samples from CIFAR-10, middle row shows the possible positions and their probabilities for each class in CIFAR-10-Position, and bottom row shows samples from all classes in CIFAR-10-Position.
}
\label{fig:data-c10-pos}
\end{figure}

We create a controlled setting to confirm that samples with known position bias score higher P-SHAP than samples with known lack of position bias. To this end we introduce the dataset \textit{CIFAR-10-Position} (see Fig.~\ref{fig:data-c10-pos}). We rescale each CIFAR-10 sample to $16 \times 16$ pixels and place it in a corner of a $32 \times 32$ pixel frame. For classes that we designate to depend on position, we always place the image in the same corner. For classes independent of position, we place the image in a random corner. CIFAR-10-Position contains eight position-dependent classes with samples $X_P$ and eight position-independent classes with samples $X_{NP}$. CIFAR-10-Position uses the same training and test split as the original CIFAR-10 dataset.

Our hypothesis is that the samples $\bX^\text{P+}$ of the position-dependent classes have a higher P-SHAP score than the samples $\bX^\text{P-}$ of the position-independent classes. To test this, we train a Vision Transformer from scratch on CIFAR-10-Position and measure P-SHAP over the test set. We use a non-parametric Mann-Whitney U-test~\cite{MannWhitney1947} with the null hypothesis that $\pshap(\bX^\text{P+}) = \pshap(\bX^\text{P-})$ and a one-sided alternative hypothesis $\pshap(\bX^\text{P+}) > \pshap(\bX^\text{P-})$. We find that the p-value of this test is below 1\%, indicating that P-SHAP can identify position-dependent samples in a model with mixed position bias. Appx.~\ref{sec:histogram} analyses the distribution of P-SHAP values within a dataset.

\subsection{P-SHAP on classification datasets}
\label{sec:position-bias-classification}

\begin{figure}
\centering
    \begin{minipage}{0.5\textwidth}
        \centering
        \includegraphics[width=0.99\linewidth]{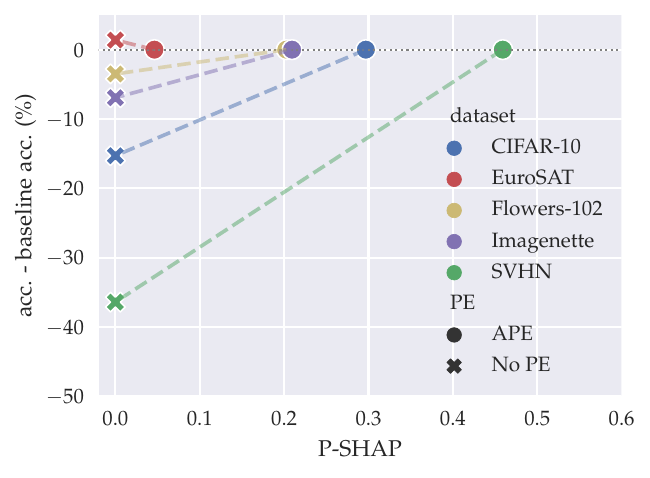}
        \captionof{subfigure}{Sec.~\ref{sec:position-bias-classification}: classification datasets}
        \label{fig:type3-datasets}
    \end{minipage}%
        \begin{minipage}{0.5\textwidth}
        \centering
        \includegraphics[width=0.99\linewidth]{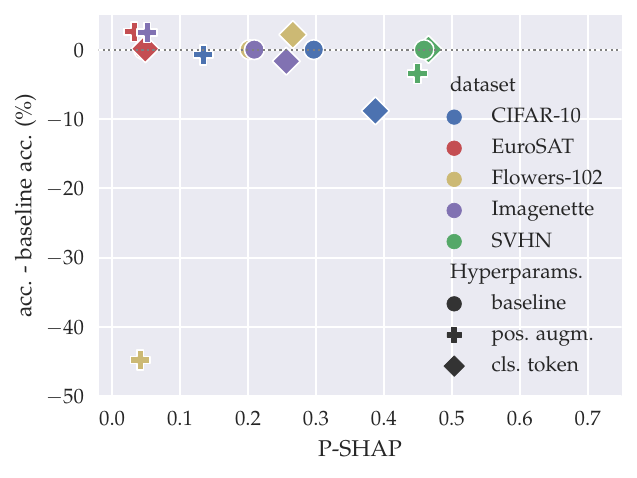}
        \captionof{subfigure}{Sec.~\ref{sec:hyperparameters}: model hyperparameters}
        \label{fig:type3-hyperparameters}
    \end{minipage}
\vspace{2mm}
\caption{Models trained from scratch on several classification datasets. Dashed lines indicate best linear fit per dataset. (a) The higher the P-SHAP in the APE model, the more performance is lost by removing the PE. (b) Position-affecting data augmentations reduce P-SHAP with predictable effects on performance, while using a class token increases P-SHAP with varying effects on performance.}
\end{figure}

To confirm that P-SHAP measures position bias in models trained on real-world datasets, we measure P-SHAP on Vision Transformers trained on two datasets chosen for their known position bias (EuroSAT and SVHN) as well as three datasets with hypothesized capture bias (CIFAR-10, Oxford-Flowers-102 and Imagenette). The training setting is detailed in Appx.~\ref{sec:training-details}.
Notably, we use a modified version of RandAugment that does not apply any position-affecting augmentations, and we use global average pooling instead of class tokens, to benefit clean experimentation w.r.t. position bias.

\textbf{Low position bias.} 
EuroSAT~\cite{helber2018introducing,helber2019eurosat} (Fig.~\ref{fig:figure-1}) is an image classification dataset of satellite imagery. The classification label pertains to a feature of the landscape in the picture, \eg a river or a highway. The imagery is translation invariant because the dataset lacks capture bias. Therefore, there is no position bias in the dataset.

\textbf{High position bias.}
SVHN~\cite{Netzer2011} (Fig.~\ref{fig:figure-1}) is an image classification dataset consisting of pictures of house numbers. While house numbers often consist of multiple digits, the images in this dataset are centered on a single digit to be classified. Therefore the network needs to distinguish between middle and other digits to learn to perform the task. Therefore, this dataset has high position bias.

\textbf{Medium position bias: capture bias.} CIFAR-10~\cite{krizhevsky2009learning}, Oxford-Flowers-102~\cite{Nilsback08} and Imagenette~\cite{imagenette}, a ten-class subset of Imagenet~\cite{imagenet} (Fig.~\ref{fig:figure-1}), all consist of natural images taken by photographers and therefore suffer from capture bias. This means that in some samples the object of interest will be in center of the image. We therefore hypothesize that P-SHAP will measure medium levels of position bias, i.e. between the levels measured for EuroSAT and SVHN.

\textbf{Results.}
Fig.~\ref{fig:type3-datasets} shows P-SHAP measured over the test set against the gain in test set accuracy compared to the baseline (APE) model. We find our hypotheses about the level of position bias are confirmed: we find that EuroSAT benefits from low P-SHAP, while SVHN benefits from high P-SHAP. Notably, EuroSAT performs better \textit{without position embeddings}, while SVHN performs much worse without position embeddings.
For the capture bias datasets, we find that all three datasets learn medium-high P-SHAP, compared to the datasets with known position bias. Correspondingly, we find that these datasets too lose performance when removing the PE, though less than for the high position bias dataset.
These performance patterns match the known position bias of the datasets, showing that position bias plays a crucial role in Vision Transformers.

\subsection{Effects of model hyperparameters}
\label{sec:hyperparameters}

% \begin{figure}
% \centering
%     \begin{minipage}{0.5\textwidth}
%     \centering
%     \includegraphics[width=0.99\linewidth]{images/type3_pshap_acc_pa.pdf}
%     \captionof{subfigure}{Sec.~\ref{sec:hyperparameters}: position-affecting data augmentations}
%     \label{fig:results-type3-pa}
%     \end{minipage}%
%     \begin{minipage}{0.5\textwidth}
%     \centering
%     \includegraphics[width=0.99\linewidth]{images/type3_pshap_acc_cls.pdf}
%     \captionof{subfigure}{Sec.~\ref{sec:hyperparameters}: class token vs. avg. pooling}
%     \label{fig:results-type3-cls}
%     \end{minipage}
% \caption{Models trained from scratch, with and without position-affecting data augmentations and class token respectively. Position-affecting data augmentations reduce P-SHAP with predictable effects on performance, while using a class token increases P-SHAP with varying effects on performance.}
% \label{fig:type3-hyperparameters}
% \end{figure}

\textbf{Position-affecting data augmentations.} All our models are trained without data augmentations that apply spatial transformations to the data. In this experiment, we compare models trained \textit{with} such augmentations to models without. Fig.~\ref{fig:type3-hyperparameters} shows that position-affecting augmentations reduce the learned P-SHAP, and that datasets with medium to high P-SHAP lose performance with position-affecting augmentations, while datasets with lower P-SHAP gain performance.
Notably, the Oxford-Flowers-102 dataset seems to be an outlier, where position augmentation almost breaks the model entirely.

\textbf{CLS token versus average pooling.} All our models are trained with average pooling instead of using a class token. Fig.~\ref{fig:type3-hyperparameters} shows that using a class token increases P-SHAP on some datasets, but performance changes unpredictably.

\textbf{Transfer learning.} We find pre-training checkpoints trained on ImageNet~\cite{imagenet} contain the roughly the same level of position bias as models trained from scratch on Imagenette, implying that models trained on more data \textit{do not} learn less position bias. 
See Appx.~\ref{sec:transferlearning} for more details and results.
% models trained with these checkpoints seem to have slightly lower position bias on the downstream task than models trained from scratch.

\subsection{Tuning for position bias}
\label{sec:tuning-position-bias}

\begin{table}[t]
\setlength{\tabcolsep}{4pt}
\small
    \centering
    \caption{Models trained from scratch to optimize for the datasets position bias. Our "tuned" setting improves accuracy significantly on low and high position bias settings we test. Scores in bold are the best in their row with a margin of one standard deviation.}
    \label{tab:tune-bias}
    \begin{tabular}{lccl}
\toprule
    \textit{dataset}  &  \textit{default (APE)} & \textit{tuned} & \textit{settings of "tuned"} \\
\midrule
    EuroSAT & 90.7 $\pm$ 0.2 & \textbf{92.4 $\pm$ 0.5} & No PE, GAP, position-affecting augm.           \\ \midrule
    CIFAR-10 & 76.5 $\pm$ 0.6 & 76.2 $\pm$ 0.2 & APE, CLS token, no position-affecting augm. \\ 
    Flowers-102 & 12.9 $\pm$ 1.4 & \textbf{60.7 $\pm$ 2.4} \\
    Imagenette & \textbf{79.1 $\pm$ 1.2} & 72.4 $\pm$ 1.4 \\  \midrule
    SVHN & 90.9 $\pm$ 0.4 & \textbf{94.8 $\pm$ 0.6} & APE, CLS token, no position-affecting augm.  \\
%     \midrule
%     & \textit{Auto-APE} & \textit{RoPE} & \textit{Auto-RoPE} \\ \midrule
%     EuroSAT &    93.3 $\pm$ 0.4 &          91.6 $\pm$ 0.2 &           93.5 $\pm$ 0.2 \\
%     CIFAR-10 &    83.5 $\pm$ 0.6 &          76.2 $\pm$ 1.1 &           84.3 $\pm$ 0.2 \\
% Flowers-102 &    58.7 $\pm$ 1.4 &          59.4 $\pm$ 0.1 &           58.4 $\pm$ 1.9 \\
%  Imagenette &    75.5 $\pm$ 1.3 &          73.1 $\pm$ 1.0 &           76.5 $\pm$ 0.3 \\
%        SVHN &    97.0 $\pm$ 0.3 &          94.7 $\pm$ 0.4 &           97.0 $\pm$ 0.2 \\
\bottomrule
\end{tabular}
\end{table}

Given the above findings, we optimize the models trained on datasets with known position bias using the settings that maximize alignment with the datasets position bias. Tab.~\ref{tab:tune-bias} shows the settings used in training as well as the results. We find that using our recommended settings significantly improves performance on the tested datasets with low and high position bias, while results are mixed on datasets with medium position bias.

\subsection{Evaluating position embeddings}
\label{sec:pe-methods}

\begin{figure}
\centering
\begin{minipage}{0.5\textwidth}
    \centering
    \includegraphics[width=0.99\linewidth]{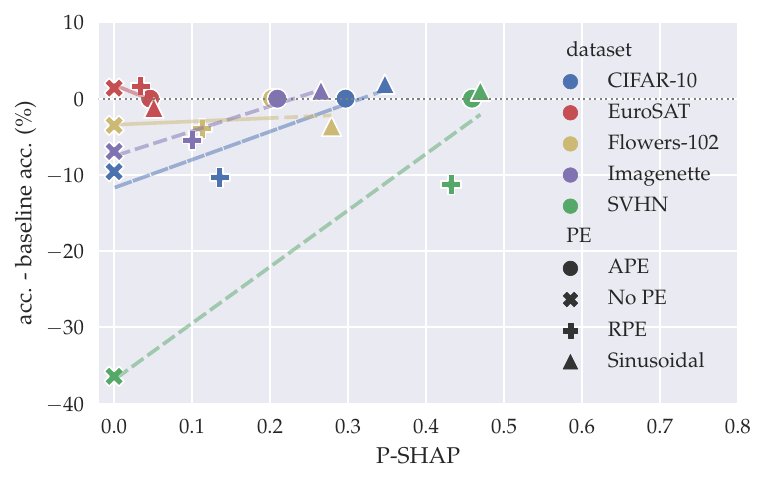}
    \captionof{subfigure}{Sec.~\ref{sec:pe-methods}: baseline PE methods}
    \label{fig:results-type3-baselinepemethods}
    \end{minipage}%
\begin{minipage}{0.5\textwidth}
    \centering
    \includegraphics[width=0.99\linewidth]{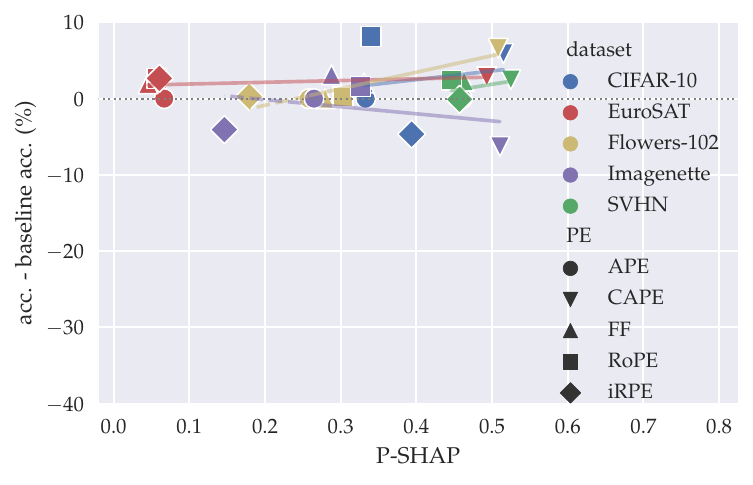}
    \captionof{subfigure}{Sec.~\ref{sec:pe-methods}: orig. contrib. PE methods}
    \label{fig:results-type3-advancedpemethods}
\end{minipage}
\vspace{2mm}
\caption{Models with varying PEs trained from scratch on several classification datasets. Dashed lines indicate best linear fit per dataset. (a) For baseline PE methods, the higher the P-SHAP in the APE model, the more performance is lost by removing the PE. (b) For orig. contrib. PE methods, there is no clear overarching trend.}
\label{fig:results-type3-pemethods}
\end{figure}

We hypothesize that the choice of position embedding is important to tune the model to the position bias of the dataset. We compare the effects on P-SHAP of using a variety of position embedding methods. We evaluate both baseline PEs
% (no PE, APE, sinusoidal and RPE)
as well as original contribution PEs.
% (iRPE, CAPE, Fourier Features (FF) and RoPE).

\textbf{Results.} Fig.~\ref{fig:results-type3-pemethods} shows the results. For the baseline methods, a similar trend as seen in the above experiments is notable: the higher the position bias in the APE model, the better the performance for models that learn more position bias. For the original contribution methods, no single overarching pattern in the relationship between position bias and accuracy emerges. 
We discuss more subtle patterns in Appx.~\ref{sec:advanced-pes}.
Crucially, RoPE improves results on all datasets
% , regardless of the position bias in the dataset.
RoPE may therefore obviate the need for a position bias-tuning PE method such as Auto-PE. However, in Sec.~\ref{sec:proposed-pes} we show that RoPE can be further improved using Auto-PE.

\subsection{Auto-PE}
\label{sec:proposed-pes}

\begin{table}[t]
\setlength{\tabcolsep}{6pt}
\small
    \centering
    \caption{Models trained from scratch with and without Auto-PE. Adding Auto-PE to either APE or RoPE equals or improves best performance on all datasets.
    % Scores in bold are the best in their row with a margin of one standard deviation.
    }
    \label{tab:proposed-pes}
    \begin{tabular}{lcccc}
\toprule
    & \multicolumn{2}{c}{\textit{APE}} & \multicolumn{2}{c}{\textit{RoPE}} \\
\cmidrule(lr){2-3}
\cmidrule(lr){4-5}
    \textit{dataset}  &  \textbf{APE} & \textbf{Auto-APE (ours)} & \textbf{RoPE} &  \textbf{Auto-RoPE (ours)} \\
\midrule
    EuroSAT & 90.7 $\pm$ 0.2          & 91.8 $\pm$ 0.3          & \textbf{93.4 $\pm$ 0.6} & \textbf{93.5 $\pm$ 0.3}              \\
   CIFAR-10 & 77.4 $\pm$ 0.7          & 77.2 $\pm$ 2.6          & 85.5 $\pm$ 0.5          & \textbf{86.3 $\pm$ 0.3}           \\
Flowers-102 & \textbf{58.6 $\pm$ 1.4} & \textbf{59.2 $\pm$ 1.0} & \textbf{58.8 $\pm$ 0.6} & 57.7 $\pm$ 0.7               \\
 Imagenette & 75.7 $\pm$ 0.7          & \textbf{76.5 $\pm$ 0.5} & \textbf{77.3 $\pm$ 2.2} & \textbf{77.7 $\pm$ 1.3}      \\
       SVHN & 94.9 $\pm$ 0.4          & 94.3 $\pm$ 0.1          & 97.3 $\pm$ 0.1          & \textbf{97.6 $\pm$ 0.1}   \\ 
\bottomrule
\end{tabular}
% &  \textbf{No PE}
% &   67.8 $\pm$ 1.7         
% &   92.1 $\pm$ 0.2
% &   55.1 $\pm$ 0.6         
% &   68.8 $\pm$ 1.1         
% &   58.5 $\pm$ 3.3

% & \multicolumn{3}{c}{\textit{PE methods (ours)}}
% & \textbf{DRAPE} &  \textbf{JAPE} &  \textbf{SCAPE}
%  & 77.6 $\pm$ 0.7 &  76.1 $\pm$ 1.0          &   76.1 $\pm$ 1.0
%  & 91.0 $\pm$ 0.5          &  90.9 $\pm$ 0.3          &   89.7 $\pm$ 0.5
%  & \textbf{58.4 $\pm$ 1.4} &  56.7 $\pm$ 0.6          &   58.0 $\pm$ 1.0
%  & 76.5 $\pm$ 1.7          &  \textbf{77.3 $\pm$ 2.2} &   \textbf{77.5 $\pm$ 1.6}
%  & 95.6 $\pm$ 0.9 &  94.3 $\pm$ 0.2          &   94.9 $\pm$ 0.5
\end{table}

% We propose Auto-PE, which allow models to learn to adapt to the position bias in the dataset.
We evaluate our proposed Auto-PE on the same datasets using the same training settings as before (see Sec.~\ref{sec:position-bias-classification}). We apply Auto-PE to APE and RoPE, the best performing baseline and original contribution PE respectively from Sec.~\ref{sec:pe-methods}.

\textbf{Results.} Table~\ref{tab:proposed-pes} shows that, even though P-SHAP is only significantly affected for some models, Auto-PE matches or outperforms the best performance on all datasets. Notably, Auto-RoPE does not significantly affect P-SHAP in any model, but still outperforms RoPE in CIFAR-10 and SVHN. An explanation can be found in Fig.~\ref{fig:results-rope-angles} in Appx.~\ref{sec:auto-pe-analysis}, which shows that Auto-PE enables RoPE to learn larger angles, leading to more discrimination between relative positions in RoPE attention. Appx.~\ref{sec:auto-pe-analysis} also contains additional results and analysis

%% file: sections/5_discussion.tex
\section{Conclusion}
\label{sec:conclusion}

In this work, we use our proposed Position-SHAP measure to show that position bias plays an important role in tuning Vision Transformers to good accuracy in image classification.
% We show that using position-related augmentations and using transfer learning affect position bias and accuracy in significant ways.
We find that the amount of position bias present in the training dataset determines the optimal choice of position embedding, and propose Auto-PE, which obviates the need for choosing a position embedding method.

\textbf{Limitations.} Position-SHAP extends Kernel SHAP,
% whose computational complexity is $\mathcal{O}(N \times B \times 2^M \times  \mathcal{O}(model))$, which makes it very slow on large \update{models/datasets}.
which is very slow on models as small as the smallest Vision Transformers (see Appx.~\ref{sec:p-shap-hyperparameters}).
In practice, hyperparameter tuning the position embedding is therefore unfortunately still more efficient than using P-SHAP. We do however see an opportunity to use DeepLIFT~\cite{shrikumar2019learningimportantfeaturespropagating} as the backbone of P-SHAP, which should make computing P-SHAP more efficient than training multiple models. Unfortunately, there is currently no implementation of DeepLIFT that is compatible with Vision Transformers.

\textbf{Future work.}
% We anticipate that P-SHAP can have several uses in future work.
P-SHAP could be used to compare pre-trained checkpoints for how appropriate they are for a specific downstream dataset, or even as backbone network for another computer vision task. We hypothesize that backbones with high learned position bias may perform better than backbones with less position bias on tasks that rely on position information, such as object detection or instance segmentation. We also anticipate that P-SHAP could be useful as a sanity check for dataset creators, ensuring that the intended position bias is indeed present, or that no position bias is accidentally infused into the data.

% https://neurips.cc/public/EthicsGuidelines
% \textbf{Broader impact.} Our work sheds light on previously unmeasurable biases in Vision Transformers. In particular, our P-SHAP method enables those applying ViTs to acknowledge and mitigate bias in their model. Otherwise, we do not foresee any direct negative societal impacts or potential for misuse arising from this research.

%% file: sections/9_appendix.tex
\section{Estimating Kernel SHAP values}
\label{sec:kernelshap}

To estimate SHAP values $\boldsymbol{\phi}^\bX$ for each sample $\bX$, an weighted linear regression problem is solved. The samples in this problem are copies of $\bX$, each with one combination of features removed from the sample. However, since neural networks do not support samples with removed features, removal of a feature is approximated by replacing it with a \textit{background value}: the mean value of the feature in a \textit{background dataset}, which should have the same distribution as the sample, \eg the training dataset. The loss function optimized in the linear regression problem minimizes the difference between the network response $f(\bX')$ on the modified sample $\bX'$ and $g(\bX')$, as shown in Theorem 2 of~\cite{LundbergLee2017}, reproduced here in simplified form:

\begin{align}
    \pi(\bX') &= \frac{(M-1)}{(M~choose~|\bX'|) |\bX'| (M - |\bX'|)} \\
    L(f,g,\pi) &= \sum_{\bX' \in \text{RemoveFeatures}(\bX)} \left[ f(\bX') - g(\bX') \right]^2 \pi(\bX'), \\
\end{align}

where $\bX \in \mathbb{R}^{HWC}$, $\text{RemoveFeatures}()$ creates one copy of $\bX$ for all possible combinations of features and replaces those features with the background values, and $|\bX'|$ is the number of non-removed features in $\bX'$. For more details, we refer to~\cite{LundbergLee2017}.

\section{Auto-RoPE}
\label{sec:auto-rope}

\textbf{RoPE.} RoPE~\cite{heo2024rotarypositionembeddingvision} is a relative position embedding method for Vision Transformers. It rotates the output of the query-key product of the attention operator in the complex domain, using (predefined but end-to-end finetuned) angles $\theta$ multiplied by the relative distance $m$ between the query and key token:

\begin{align}
    % f_q(\x_m, m) &= (\W_q\x_m)e^{im\theta}\\
    % f_k(\x_n, n) &= (\W_k\x_n)e^{in\theta}\\
    g(\x_m, \x_n, m-n) &= \mathrm{Re}[(\W_q \x_m) (\W_k \x_n)^* e^{i(m-n)\theta}]
\end{align}

where $g$ computes RoPE attention for input tokens $\x_m$ and $\x_n$, that are $m-n$ positions apart, by multiplying the query-key product with complex-valued $e^{i(m-n)\theta}$, which amounts to rotation in the complex domain by a factor dependent on (learnable) parameter $\theta$ and the relative position difference $m-n$. This is the computation of RoPE for a single combination of query and key tokens. For more details, see~\cite{heo2024rotarypositionembeddingvision} for the original RoPE formulation and~\cite{heo2024rotarypositionembeddingvision} for the Vision Transformer formulation.

The angles $\theta$ are initialized as they are in the sinusoidal APE encoding~\cite{vaswani2017attentionneed}, but they are fine-tuned in an end-to-end manner during training. This fine-tuning allows the model to "unlearn" the relative position embedding that RoPE applies, by learning $\theta = 0$. The results in Sec.~\ref{sec:pe-methods} suggest that RoPE is reasonably successful in this. However, we proceed to implement Auto-RoPE and show in Sec.~\ref{sec:proposed-pes} that further improvement is possible.

\textbf{Auto-RoPE.} We design a special case of Auto-PE for RoPE~\cite{heo2024rotarypositionembeddingvision}, implementing the position embedding modulation of Auto-PE in RoPE by applying a learned weight $\gamma$ to $\theta$:

\begin{align}
    f_q(\x_m, m) &= (\W_q\x_m)e^{im\gamma\theta}\\
    f_k(\x_n, n) &= (\W_k\x_n)e^{in\gamma\theta}.\\
\end{align}

Auto-RoPE modulates the angle of rotation that RoPE applies. When $\gamma = 0$, the effective angle of rotation $\theta' = \gamma \theta = 0$.
% \update{We learn a single parameter $\gamma$ for the whole network, which means that $\gamma$ is shared between all attention operators in the network.}
We learn a single parameter $\gamma$ for each attention operator in the network.
% \todo{Update all results with RoPE with shared $\gamma$.}
% We learn a single parameter $\gamma$ for every RoPE attention operator, which means that $\gamma$ is shared between all relative position offsets and heads of an attention operator, but not between attention operators in the network.

\section{Validity of Kernel SHAP on ViT}
\label{sec:slalom}

SLALOM~\cite{leemann2025attention} argues that model linearity is an incorrect assumption for Vision Transformers (ViTs), and therefore Kernel SHAP should not be used on ViTs. Their theoretical argument is based on the assumption that identical tokens can exist in the input to the transformer model. This is indeed the case when using ViTs on text, as the same word appearing more than once in a sentence will be represented by identical tokens. However, because Vision Transformers take linearly projected patches as input tokens to which further position embeddings are added, we argue the likelihood of encountering identical tokens is negligible. Additionally, the empirical results in~\cite{leemann2025attention} suggest that Kernel SHAP performs well in non-synthetic datasets. We therefore argue that SHAP is a valid feature attribution method for Vision Transformers, and use it in this sense in our work.

\section{Analysis of P-SHAP values of individual samples}
\label{sec:histogram}

\begin{figure}
\centering
\includegraphics[width=0.5\linewidth]{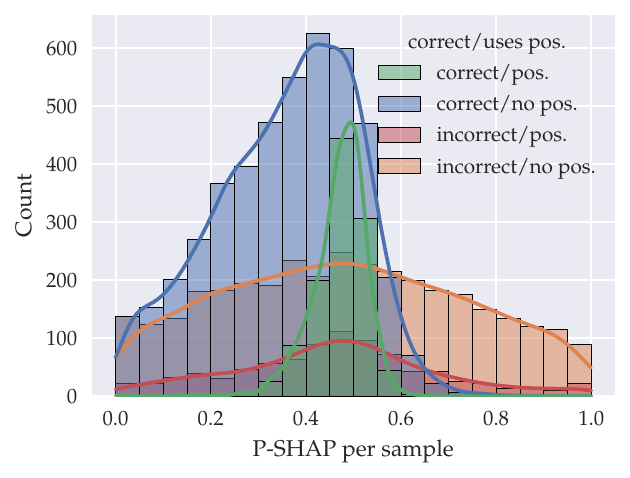}
\caption{
Histogram of CIFAR-10-Position validation samples, split by correct/incorrect prediction and position in- or dependent. Samples independent of position still use position embeddings to link neighboring patches, but P-SHAP is on average higher in position-dependent samples.
}
\label{fig:c10-pos-histogram}
% \vspace{-3mm}
\end{figure}

Figure~\ref{fig:c10-pos-histogram} shows a histogram of P-SHAP values of all samples in the test set for the experiment in Sec.~\ref{sec:type1}, split by correctly and incorrectly classified samples. This shows the test passes because the means of the distributions of the correctly classified samples are clearly different, but still there is some overlap between the distributions. In any model, we will measure some position bias for every sample, since position embeddings are necessary to link neighboring image patches regardless of their position. Still, we measure more position bias for samples dependent on the position of the object of interest.

Furthermore, there is a clear difference between the P-SHAP values of correctly classified samples and the incorrectly classified samples. To reduce noise in the P-SHAP measurement, we only measure P-SHAP for correctly classified samples, before taking the mean value to be the position bias of the trained model.

\section{Effects of P-SHAP hyperparameters}
\label{sec:p-shap-hyperparameters}

\paragraph{P-SHAP variance.} We run P-SHAP six times on the same trained model checkpoint to compute the standard deviation induced from the randomness in P-SHAP alone. This standard deviation is $0.0039$ for a model that scores a mean P-SHAP of $0.3854$.

\paragraph{Batch size.} We compute P-SHAP by taking half of the batch as background to compute the P-SHAP values of the other half of the batch, and vice versa. The size of the batch therefore has an effect on the P-SHAP estimate, as it determines the size of the background dataset, which will be a better estimate of the true background with more samples. In our experiments the default batch size is 32 samples. We conduct an experiment finding that a larger batch size increases P-SHAP inference time notably without substantial effect on the variance in the P-SHAP estimate.

\paragraph{Inference cost.} P-SHAP is on average $5571$ times slower than regular validation inference: on CIFAR-10-Position, running the model once on the validation set takes approx. eight seconds, while running P-SHAP takes on average 11 hours and 12 minutes. We address this limitation further in Sec.~\ref{sec:conclusion}.

\section{Analysis of P-SHAP on advanced PE methods}
\label{sec:advanced-pes}

\begin{figure}
    \centering
    \includegraphics[width=0.79\linewidth]{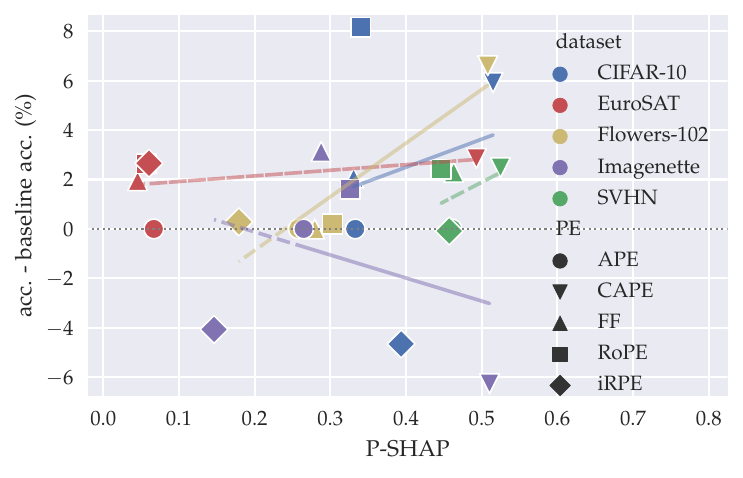}
    \vspace{2mm}
    \caption{Reproduction of Fig.~\ref{fig:results-type3-advancedpemethods} in Sec.~\ref{sec:pe-methods}: advanced PE methods}
    \label{fig:advanced-pes-appx}
\end{figure}

We analyze the results of Sec.~\ref{sec:pe-methods} in more detail. See Fig.~\ref{fig:advanced-pes-appx} for a large reproduction of Fig.~\ref{fig:results-type3-advancedpemethods}.

CAPE increases P-SHAP the most of all advanced PE methods, with strongly varying effect on accuracy: it performs worse on ImageNette, but (on par with) best on the other datasets.
FF and RoPE do not increase P-SHAP significantly, but do increase performance on almost all datasets. This indicates the inductive bias of a PE can contribute to better performance in a different way than merely increasing the learned position bias.
Finally, iRPE at best does not improve performance, and at worse reduces performance by a large margin.
We do note that on EuroSAT, the dataset with the lowest position bias, the earlier observed pattern that learning more position bias does not help, can be noted here as well. Similarly, on SVHN, the dataset with the highest position bias, all methods learn high position bias, though the slight performance differences we see cannot be directly attributed to higher learned position bias. We conclude that the benefits of advanced PE methods are not mainly in how they increase position bias, but rather may be in the way in which they allow models to process positional information.

\section{Training details}
\label{sec:training-details}

\paragraph{Default setting (Sec.~\ref{sec:position-bias-classification},~\ref{sec:hyperparameters},~\ref{sec:tuning-position-bias},~\ref{sec:pe-methods} and~\ref{sec:proposed-pes}).} We adapt the setup of~\cite{gani2022trainvisiontransformersmallscale}: we use ViT models~\cite{dosovitskiy2021imageworth16x16words} with 4px patch size, 192 dimensions in the hidden layer, 384 dimensions in the MLP, 12 heads and 9 layers, applying dropout with magnitude 0.1, training from scratch while downsampling all datasets to 32px image resolution. For data augmentations, we MixUp~\cite{zhang2018mixup}, CutMix~\cite{yun2019cutmix}, random erasing~\cite{zhong2020random} and RandAugment~\cite{cubuk2020randaugment}. However, we exclude RandAugment operations that affect the image spatially (e.g. rotation, scaling, shearing, etc.), as we perform a separate experiment in Sec.~\ref{sec:hyperparameters} on to test the effects of such data augmentations. We train all models for 400 epochs with batch size 512, using an Adam optimizer~\cite{kingma2017adammethodstochasticoptimization} with learning rate $1\mathrm{e}{-3}$ decayed using cosine annealing~\cite{loshchilov2017sgdrstochasticgradientdescent}, and label smoothing with magnitude $0.1$. The models using APE position embedding are treated as the default models in our experiments, and perform comparable to published results with similar model sizes~\cite{dosovitskiy2021imageworth16x16words}.
We train each model three times and average the results.

\paragraph{Transfer learning setting (Appx.~\ref{sec:transferlearning}).} We use the same setup as in Sec.~\ref{sec:position-bias-classification}, except we change the model to use the original ViT implementation~\cite{dosovitskiy2021imageworth16x16words} of the \texttt{ViT-Tiny/16} model, with pretrained weights as published on Huggingface~\cite{huggingface}\footnote{Weights from \url{https://huggingface.co/WinKawaks/vit-tiny-patch16-224/tree/main}}. This model is larger than the model we used in previous experiments, but the smallest ViT model that we could find and successfully fits to the downstream datasets, even training from scratch. We train each model only once, to save compute.

\section{Effects of transfer learning}
\label{sec:transferlearning}

\begin{figure}
\centering
    % \begin{minipage}{0.5\textwidth}
    % \centering
    \includegraphics[width=0.5\linewidth]{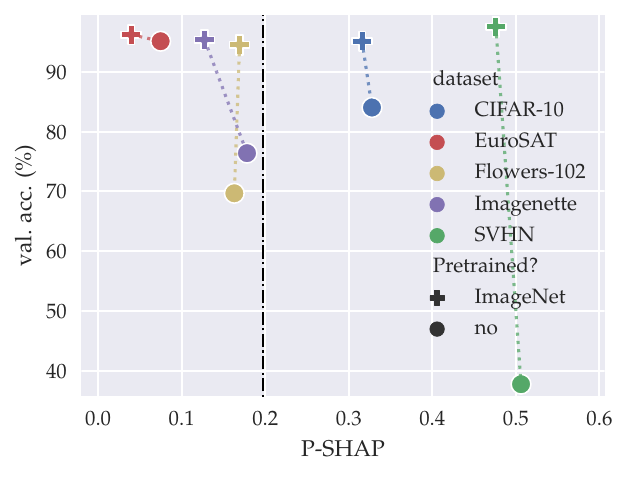}
    % \captionof{subfigure}{Sec.~\ref{sec:transferlearning}: transfer learning}
    % \label{fig:results-type3-transferlearning}
    % \end{minipage}%
% \vspace{2mm}
\caption{Results for models trained with and without pre-trained weights. Dash-dot line indicates the P-SHAP of the pre-trained checkpoint. We find that transfer learning affects position bias, slightly reducing it. Datasets with high position bias lose more accuracy by leaving out fine-tuning than datasets with low position bias.
}
\label{fig:results-type3-transferlearning}
\end{figure}

For transfer learning to be successful, the pre-trained model checkpoints should either be free of any dataset-specific bias or this bias needs to match the downstream dataset. We test if P-SHAP is learned during pre-training by measuring P-SHAP in the checkpoint. We then compare P-SHAP in models trained from scratch and with transfer learning on the aforementioned datasets. The training setting is detailed in Appx.~\ref{sec:training-details}.

\textbf{Position bias in pre-trained model.} We compute P-SHAP for the checkpoint on a random subset of 43.1\% of the ImageNet validation set. Fig.~\ref{fig:results-type3-transferlearning} shows that this model scores a P-SHAP of $\approx 0.2$, which we consider to be a medium-low amount of position bias. Interestingly, this score is not zero or even as low as learned on the low position bias dataset EuroSAT, implying that ImageNet pre-training does learn some position bias that is passed on to the downstream models.
Furthermore, the measured position bias roughly matches the measured position bias of the related dataset Imagenette, which has only ten classes and $100\times$ less samples. This implies that models trained on more data \textit{do not} learn less position bias.

\textbf{Accuracy in downstream models.} As expected, models using transfer learning perform significantly better than models trained from scratch. Interestingly, the higher the position bias, the bigger the gap in accuracy between fine-tuning and training from scratch. This could indicate that position bias actually hinders models trained from scratch more than those that are fine-tuned. We do note that CIFAR-10 seems to be a slight outlier in this trend.

\textbf{Position bias in downstream models.} Notably, position bias seems to be slightly smaller in fine-tuned models than in models trained from scratch. We speculate this could be because of the model needing to re-adjust its learned position bias to the position bias of the downstream dataset, which may be difficult, and lead to the model not being able to use position bias as well.

\section{Analysis of Auto-PE results}
\label{sec:auto-pe-analysis}

\begin{table}[t]
\setlength{\tabcolsep}{6pt}
\small
    \centering
    \caption{Combining the tuning for position bias with Auto-PE: models trained from scratch with and without Auto-PE, on the baseline and "tuned" setting from Sec.~\ref{sec:tuning-position-bias}. Combining our recommended "tuned" setting with Auto-PE does not lead to further improved results. Scores in bold are the best in their row with a margin of one standard deviation.}
    \label{tab:proposed-pes-full}
    \scalebox{0.75}{
    \begin{tabular}{lcccccccc}
\toprule
    & \multicolumn{2}{c}{\textit{standard setting}} & \multicolumn{2}{c}{\textit{+ Auto-PE}} & \multicolumn{2}{c}{\textit{+ tuned}} & \multicolumn{2}{c}{\textit{+ tuned + Auto-PE}} \\
\cmidrule(lr){2-3}
\cmidrule(lr){4-5}
\cmidrule(lr){6-7}
\cmidrule(lr){8-9}
    \textit{dataset}  &  \textbf{APE} & \textbf{RoPE} &  \textbf{Auto-APE} & \textbf{Auto-RoPE} &  \textbf{APE/No PE} & \textbf{RoPE} & \textbf{Auto-APE} & \textbf{Auto-RoPE} \\
\midrule
    EuroSAT & 90.8 $\pm$ 0.3 & 92.2 $\pm$ 0.2          & 92.0 $\pm$ 0.7 & 92.3 $\pm$  0.5          & 92.4 $\pm$ 0.6          & \textbf{93.3 $\pm$ 0.4} & 91.6 $\pm$ 0.2          & 93.5 $\pm$ 0.2              \\
   CIFAR-10 & 76.5 $\pm$ 0.6 & \textbf{88.1 $\pm$ 0.3} & 75.9 $\pm$ 0.2 & 86.5 $\pm$  1.1          & 76.2 $\pm$ 0.2          & 83.5 $\pm$ 0.6          & 76.2 $\pm$ 1.1          & 84.3 $\pm$ 0.2            \\
Flowers-102 & 12.9 $\pm$ 1.4 & 13.4 $\pm$ 0.6          & 14.0 $\pm$ 0.9 & 13.6 $\pm$  0.5          & \textbf{60.7 $\pm$ 2.4} & \textbf{58.7 $\pm$ 1.4} & \textbf{59.4 $\pm$ 0.1} & 58.4 $\pm$ 1.9              \\
 Imagenette & 79.1 $\pm$ 1.2 & \textbf{81.7 $\pm$ 1.0} & 79.5 $\pm$ 1.5 & \textbf{80.9 $\pm$  0.7} & 72.4 $\pm$ 1.4          & 75.5 $\pm$ 1.3          & 73.1 $\pm$ 1.0          & 76.5 $\pm$ 0.3              \\
       SVHN & 90.9 $\pm$ 0.4 & 93.4 $\pm$ 1.1          & 90.6 $\pm$ 0.3 & 86.3 $\pm$ 10.4          & 94.8 $\pm$ 0.7          & \textbf{97.0 $\pm$ 0.3} & 94.7 $\pm$ 0.4          & \textbf{97.0 $\pm$ 0.2}                   \\ 

\bottomrule
\end{tabular}}
\end{table}

\begin{figure}
\centering
    \begin{minipage}{0.5\textwidth}
    \centering
    \includegraphics[width=0.99\linewidth]{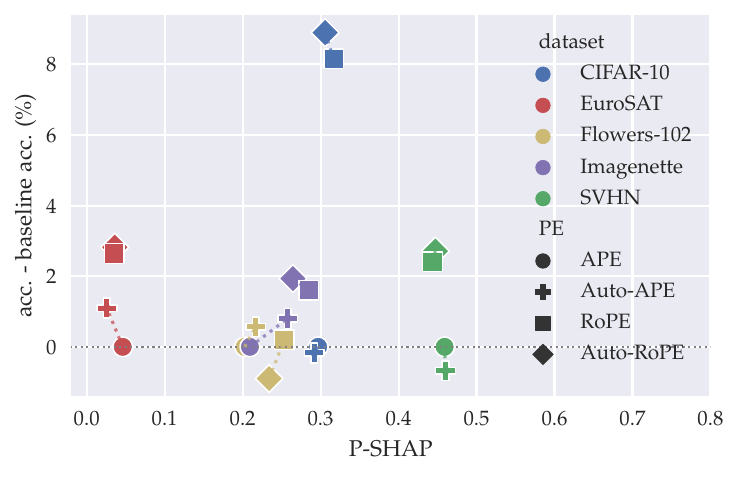}
    \captionof{subfigure}{Sec.~\ref{sec:proposed-pes}: APE and RoPE with Auto-PE}
    \label{fig:results-type3-autope-shap}
    \end{minipage}%
    % \begin{minipage}{0.5\textwidth}
    % \centering
    % \includegraphics[width=0.99\linewidth]{images/rope_angles.pdf}
    % \captionof{subfigure}{Sec.~\ref{sec:proposed-pes}: Auto-RoPE learned angles}
    % \label{fig:results-rope-angles}
    % \end{minipage}
\vspace{3mm}
\caption{Results for Auto-PE. Dotted lines link models with and without Auto-PE. Auto-PE matches or improves the baseline PE in performance by learning the same or more dataset-appropriate position bias, as measured by P-SHAP.}
\label{fig:type3-autope}
\end{figure}

\begin{figure}
\centering
    \begin{minipage}{0.5\textwidth}
    \centering
    \includegraphics[width=0.99\linewidth]{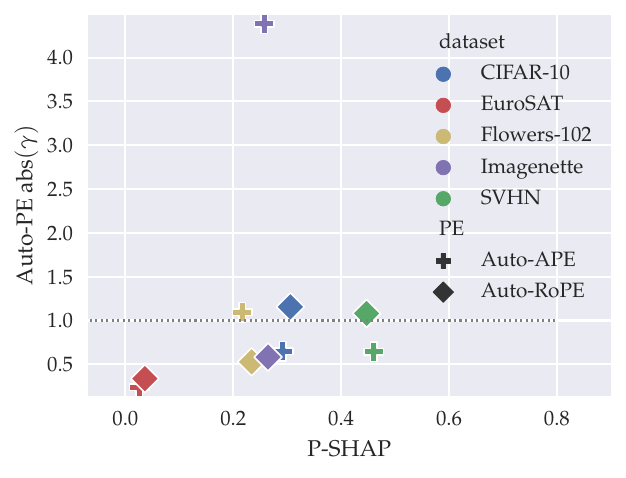}
    \captionof{subfigure}{Sec.~\ref{sec:proposed-pes}: Auto-PE: $\mathrm{abs}(\gamma)$}
    \label{fig:results-type3-autope-shap}
    \end{minipage}%
    \begin{minipage}{0.5\textwidth}
    \centering
    \includegraphics[width=0.99\linewidth]{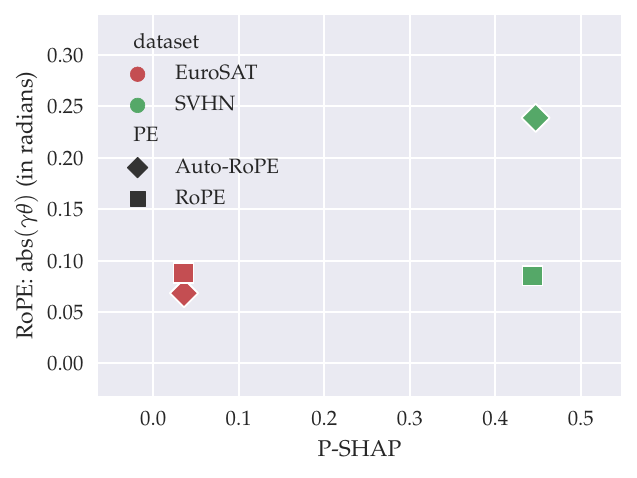}
    \captionof{subfigure}{Sec.~\ref{sec:proposed-pes}: Auto-RoPE learned angles}
    \label{fig:results-rope-angles}
    \end{minipage}
\vspace{2mm}
\caption{Learned values of (a) $\gamma$ of Auto-APE and Auto-RoPE and (b) angles of Auto-RoPE. Dotted lines link models with and without Auto-PE. Auto-PE nearly always learns to reduce the norm of the PE, though this effect is larger on datasets with low position bias. In Auto-RoPE, Auto-PE allows RoPE's learned angles to converge to larger values, though we do not measure higher P-SHAP in these models. The $\gamma$-values for Auto-RoPE are the mean value of all $\gamma$ parameters in the network.}
\label{fig:type3-autope-learned}
\end{figure}

Fig.~\ref{fig:type3-autope} shows the results of Auto-PE plotted against position bias. Auto-PE does not seem to significantly alter position bias in any consistent manner, compared to models trained without Auto-PE. One exception is the Auto-APE model on EuroSAT, which we know from Sec.~\ref{sec:position-bias-classification} needs to unlearn the PE. This model learns less position bias and improved accuracy, like when using no PE.

Fig.~\ref{fig:results-type3-autope-shap} shows the absolute value of learned $\gamma$ parameters in Auto-PE. We note that learned values of $\gamma$ are sometimes negative, thereby inverting the values of the PE. Since this should not have an effect on the position bias modeled, we plot the absolute value of the $\gamma$ parameters. We note that Auto-PE nearly always learns to reduce the norm of the PE, though this effect is larger on datasets with low position bias.

Fig.~\ref{fig:results-rope-angles} shows the effective rotation angles applied in Auto-RoPE attention, after the angles $\theta$ have been modulated by $\gamma$. Notably, on EuroSAT, Auto-RoPE reduces the effective angle, attenuating the position bias, while on SVHN the effective angle is increased, increasing the position bias as well.

\textbf{Additional results.} Preliminary experiments showed that Auto-RoPE with a single parameter $\gamma$ shared between all attention operators performed slightly worse. We therefore did not consider this implementation in our work any further.